\documentclass[11pt,letterpaper]{article}
\usepackage{naaclhlt2016}
\usepackage{times}
\usepackage{latexsym}
\usepackage[usenames,dvipsnames]{color}
\usepackage{url}
\usepackage{lipsum}

\usepackage{graphicx}
\usepackage{amsmath,amssymb}
\usepackage{subfig}
\usepackage{tablefootnote}
\usepackage{bm}
\usepackage{booktabs}
\usepackage{siunitx}
\usepackage{xspace}
\usepackage{xcolor}
\usepackage{soul}
\usepackage[warn]{textcomp}
\usepackage{paralist}

\newcommand{\sos}{\textlangle s\textrangle\xspace}
\newcommand{\eos}{\textlangle /s\textrangle\xspace}
\newcommand{\unk}{\textlangle unk\textrangle\xspace}

\renewcommand{\vec}[1]{\bm{#1}}

\newcommand{\mat}[1]{\mathbf{#1}}
\newcommand{\vsuper}[2]{\vec{#1}^{(\text{#2})}}
\newcommand{\msuper}[2]{\mat{#1}^{(\text{#2})}}

\newcommand\vs{\vec{s}}
\newcommand\vt{\vec{t}}

\newcommand\ve{\vec{e}}

\newcommand\vf{\vec{f}}
\newcommand\vh{\vec{h}}

\newcommand\vu{\vec{u}}
\newcommand\vv{\vec{v}}

\newcommand\valpha{\vec{\alpha}}
\newcommand\vhf{\vec{h}^{\rightarrow}}
\newcommand\vhb{\vec{h}^{\leftarrow}}
\newcommand\Rs{\msuper{R}{s}}
\newcommand\rs{\vsuper{r}{s}}

\newcommand\Wsif{\vec{W}_{si}^{\rightarrow}}
\newcommand\Wsib{\vec{W}_{si}^{\leftarrow}}
\newcommand\Wshf{\vec{W}_{sh}^{\rightarrow}}
\newcommand\Wshb{\vec{W}_{sh}^{\leftarrow}}
\newcommand\bsf{\vec{b}_{s}^{\rightarrow}}
\newcommand\bsb{\vec{b}_{s}^{\leftarrow}}
\newcommand\vg{\vec{g}}

\newcommand\rt{\vsuper{r}{t}}
\newcommand\Wti{\msuper{W}{ti}}
\newcommand\Wth{\msuper{W}{th}}

\newcommand\bto{\vsuper{b}{to}}

\newcommand\sizev[1]{\in \mathbb{R}^{#1}}
\newcommand\sizem[2]{\in \mathbb{R}^{#1 \times #2}}

\newcommand\softmax{\operatorname{softmax}}

\naaclfinalcopy 
\def\naaclpaperid{***} 

\ifnaaclfinal
\fi

\newcommand\vE{\vec{e}}

\title{Incorporating Structural Alignment Biases into an Attentional Neural Translation Model}

\author{Trevor Cohn \and Cong Duy Vu Hoang \and Ekaterina Vymolova \\
	    University of Melbourne\\
	    Melbourne, VIC, Australia\\
	    {\tt tcohn@unimelb.edu.au} \and
	    {\tt \{vhoang2,evylomova\}@student.unimelb.edu.au} 
	  \AND
	Kaisheng Yao\\
  	Microsoft Research\\
  	Redmond, WA, USA\\
        {\tt kaisheng.YAO@microsoft.com}~~~
	  \And
	Chris Dyer\\
  	Carnegie Mellon University\\
  	Pittsburgh, PA, USA\\
        {\tt cdyer@cmu.edu} 
          \And
	Gholamreza Haffari\\
  	Monash University\\
  	Clayton, VIC, Australia\\
        ~~~{\tt gholamreza.haffari@monash.edu}
}

\date{}

\begin{document}

\maketitle

\begin{abstract}
Neural encoder-decoder models of machine translation have achieved impressive results,
rivalling traditional translation models. However their modelling formulation
is overly simplistic, and omits several key inductive biases built into
traditional models. In this paper we extend the attentional neural 
translation model to include structural biases from word based alignment models, 
including positional bias, Markov conditioning, fertility and agreement over translation directions. 
We show improvements over a baseline attentional model and standard phrase-based model 
over several language pairs, evaluating on difficult languages in a low resource setting.%
\end{abstract}

\section{Introduction}
Recently, models of end-to-end machine translation based
on neural network classification have been shown to produce excellent
translations, rivalling or in some cases surpassing traditional statistical machine
translation systems \cite{kalchbrenner13emnlp,sutskever2014sequence,bahdanau2015neural}.
This is despite the neural approaches using an overall simpler model, with fewer assumptions
about the learning and prediction problem.

Broadly, neural approaches are based around the notion of an
\emph{encoder-decoder}
\cite{sutskever2014sequence}, in which the source language is \emph{encoded}
into a distributed representation, followed by a \emph{decoding}
step which generates the target translation. 
We focus on the \emph{attentional model} of translation \cite{bahdanau2015neural}  
which uses a dynamic representation of the source sentence 
while allowing the decoder to \emph{attend} to different parts of
the source as it generates the target sentence.
The attentional model raises intriguing opportunities, given the correspondence between the notions of attention 
and alignment in traditional word-based  machine translation models \cite{brown93}.

In this paper we map modelling biases from word based translation models into the attentional model,
such that known linguistic elements of translation can be better captured. 
%
%
We incorporate \emph{absolute positional bias} whereby word order 
tends to be similar between the source sentence and its translation 
(e.g., IBM Model 2 and  \cite{dyer-chahuneau-smith:2013:NAACL-HLT}), 
 \emph{fertility} whereby each instance of a source word type tends 
to be translated into a consistent number of target tokens (e.g., IBM Models 3, 4, 5), 
\emph{relative position bias} whereby prior preferences for monotonic 
alignments/attention can be encouraged  (e.g., IBM Model 4, 5 and HMM-based Alignment \cite{vogel96}),
and \emph{alignment consistency} whereby the attention in \emph{both} translation 
directions are encourged to agree  (e.g. symmetrization heuristics \cite{och03} 
or joint modelling \cite{liang2006alignment,ganchev-gracca-taskar:2008:ACLMain}). 

We provide an empirical analysis of incorporating the above structural biases into the attentional model, 
considering low resource translation scenario over four language-pairs. 
Our results demonstrate consistent improvements over vanila encoder-decoder and attentional model in 
terms of the perplexity and BLEU score, e.g.\@ up to 3.5 BLEU points   
when re-ranking the candiate translations generated by a state-of-the-art phrase based model.%
\ifnaaclfinal\else\footnote{The source code will be released on publication}\fi



\section{The attentional model of translation}
\label{sec:sentence-encdec}

\begin{figure}[t]
\centering
\includegraphics[width=0.8\columnwidth]{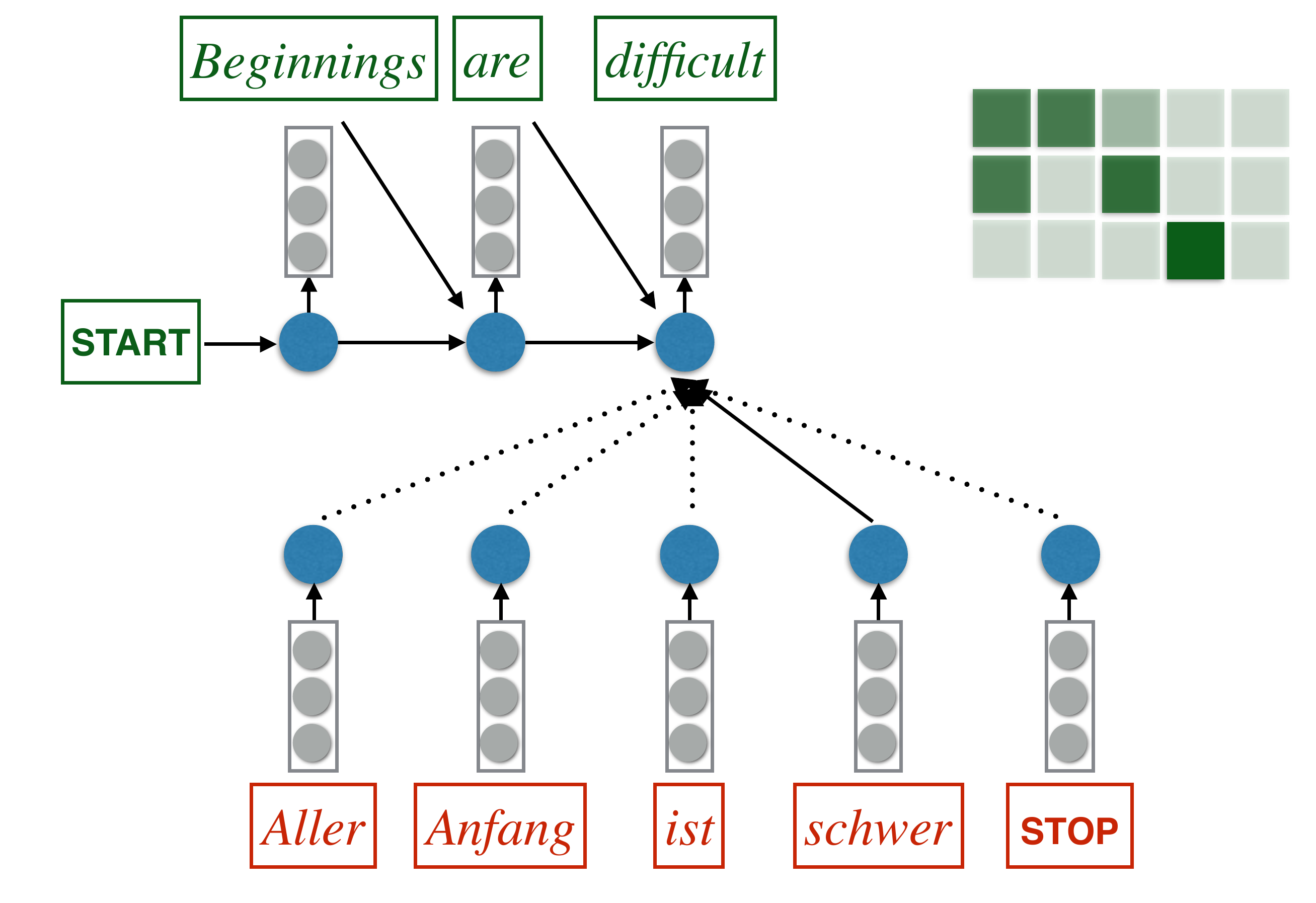}
\caption{Attentional model of translation
  \protect\cite{bahdanau2015neural}. The encoder is shown below the decoder,
  and the edges connecting the two corresponding to the attention
  mechanism. Heavy edges denote a higher attention weight, and these
  values are also displayed in matrix form, with one row for each target word. }
\label{fig:sentence-attentional}
\end{figure}

We start by reviewing the attentional model of translation \cite{bahdanau2015neural},
as illustrated in Fig.~\ref{fig:sentence-attentional}, before presenting our extensions in \S\ref{sec:sent-alignment-structure}.

\newcommand\E{\mat{E}}

\paragraph{Encoder} The encoding of the source sentence is formulated
using a pair of RNNs (denoted \emph{bi-RNN}) one operating
left-to-right over the input sequence and another operating
right-to-left,
\begin{align*}
\vhf_i &= \text{RNN}(\vhf_{i-1}, \rs_{s_i}) \\
\vhb_i &= \text{RNN}(\vhf_{i+1}, \rs_{s_i}) 
\end{align*}
where $\vhf_i$ and $\vhb_i$ are the RNN hidden states.  
The left-to-right RNN
function is defined as 
\begin{align}
\vhf_i = \tanh\left( \Wsif \rs_{s_i} +  \Wshf \vhf_{i-1} +  \bsf \right)
\label{eq:sent-rnn-encoder}
\end{align}
where $\vh_0^{\rightarrow} \sizev{H}$ is a learned parameter vector, as are
$\Rs \sizem{V_S}{E}$,
$\Wsif \sizem{H}{E}$,  $\Wshf \sizem{H}{H}$ and $\bsf \sizev{H}$, with
$H$ the number of hidden units,  $V_S$ the size of the source
vocabulary and $E$ the word embedding
dimensionality.\footnote{Similarly, $\vh_0^{\leftarrow} \sizev{H}, \Wsib \sizem{H}{E}, \Wshb \sizem{H}{H}, \bsb \sizev{H}$ are the parameters of the right-to-left RNN. Note that we use a long short term memory unit \cite{Hochreiter1997} in place of the RNN, shown here for simplicity of exposition.}
Each source word is then
 represented as a pair of hidden states, one from each RNN, 
$\vE_i  = \left[ \begin{array}{c} \vhf_i \\ \vhb_i \end{array} \right]$.
This encodes not only the word but also its left and right context,
which can provide important evidence for its translation.

A crucial question is how this dynamic sized matrix 
$\E = \left[ \vE_1, \vE_2, \ldots, \vE_I\right] \sizem{I}{H}$ 
can be used in the decoder to generate the target sentence.
As with Sutskever's encoder-decoder, the target sentence is created
left-to-right using a RNN, while the encoded source is used to bias
the process as an auxiliary input. The mechanism for this
bias is by attentional vectors, i.e. vectors of scores
over each source sentence location, which are used to aggregate the
dynamic source encoding into a fixed length vector.

\newcommand{\vc}{\vec{c}}
\newcommand{\Wta}{\msuper{W}{ta}}
\newcommand{\Wuc}{\msuper{W}{uc}}
\newcommand{\Wui}{\msuper{W}{ui}}
\newcommand{\Wou}{\msuper{W}{ou}}

\paragraph{Decoder} The decoder operates as a standard RNN over the  translation
$\vt$, formulated as follows
{\small
\begin{align}
\vg_{j} & = \tanh \left( \Wth \vg_{j-1} + \Wti \rt_{t_{j-1}} + \Wta \vc_j \right) 
\label{eq:sent-am-decoder} \\
\vu_j & = \tanh \left( \vg_j +  \Wuc \vc_j + \Wui \rt_{t_{j-1}}  \right) 
\label{eq:sent-am-hidden-out} \\
t_{j} & \sim \softmax \left( \Wou \vu_j + \bto  \right)
\label{eq:sent-am-out}
\end{align}
}%
where the decoder RNN is defined analogously to
Eq~\ref{eq:sent-rnn-encoder} but with an additional input, the source attention component
$\vc_j \sizev{2H}$ and weighting matrix $\Wta \sizem{H}{2H}$.
The hidden state of the recurrence is then passed through a single
hidden layer\footnote{In \newcite{bahdanau2015neural} they use a max-out
  layer for this final step, however we found this to be a needless
  complication, and instead use a standard hidden layer with tanh activation.}
 (Eq~\ref{eq:sent-am-hidden-out}) in combination with the
 source attention and target word 
using weighting matrices
$\Wuc \sizem{H}{2H}$ and
$\Wui \sizem{H}{E}$.
In Eq~\ref{eq:sent-am-out} this vector is transformed to be
target vocabulary sized, using weight matrix 
$\Wou \sizem{V_T}{H}$ and bias $\bto \sizev{V_T}$, after which a $\operatorname{softmax}$ is taken, and the resulting normalised
vector used as the parameters of a Categorical distribution in generating the next target word.

The presentation above assumes a simple RNN is used to define the
recurrence over hidden states, however we can easily use alternative
formulations of recurrent networks including multiple-layer RNNs,
gated recurrent units (GRU; \newcite{ChoGRU2014}), or long short-term memory (LSTM; \newcite{Hochreiter1997}) units.
These more advanced methods allow for more efficient learning of more
complex concepts, particularly long distance effects. Empirically we found
LSTMs to be the best performing, and therefore use these units herein.

\newcommand\Wae{\msuper{W}{ae}}
\newcommand\Wac{\msuper{W}{ac}}
\newcommand\Wah{\msuper{W}{ah}}

The last key detail is the attentional component $\vc_j$ in Eqs~\ref{eq:sent-am-decoder} and~\ref{eq:sent-am-hidden-out}, which is defined as follows
\begin{align}
f_{ji} & = \vv^\top \tanh \left( \Wae \vE_i + \Wah \vg_{j-1}\right) \label{eq:sent-fji} \\
\valpha_j & = \softmax \left( \vf_j \right) \nonumber \\ 
\vc_j &= \sum_i \alpha_{ji} \vE_i \nonumber
\end{align}
with the scalars $f_{ji}$ denoting the compatibility between the target hidden state
$\vg_{j-1}$ and the source encoding $\vE_i$. This is defined as a neural
network with one hidden layer of size $A$ and a single output, parameterised by
$\Wae \sizem{A}{2H}$,
$\Wah \sizem{A}{H}$ and
$\vv \sizev{A}$.
The $\operatorname{softmax}$ then normalises the scalar compatibility values such that
for a given target word $j$, the values of $\alpha_j$ can be
interpreted as alignment probabilities to each source
location. Finally, these alignments are used to to reweight
the source components $E$ to produce a fixed length context 
representation.

Training of this model is done by minimising the cross-entropy of the target sentence,
measured word-by-word as for a language model. 
We use standard stochastic gradient
optimisation using the back-propagation technique for computation of
partial derivatives according to the chain rule.


\section{Incorporating Structural Biases}
\label{sec:sent-alignment-structure}

The attentional model, as described above, provides a powerful and
elegant model of translation in which alignments between source and
target words are learned through the implicit conditioning context afforded
by the attention mechanism. Despite its elegance, the attentional model omits several key 
components of a traditional alignment models such as the IBM models
\cite{brown93} and Vogel's hidden Markov Model \cite{vogel96} as
implemented in the GIZA++ toolkit \cite{och03}. Combining the strengths
of this highly successful body of research into a neural model of
machine translation holds potential to further improve modelling
accuracy of neural techniques. 
Below we outline methods for incorporating these factors as structural biases into the attentional model.

\subsection{Position bias}

\newcommand{\Wap}{\msuper{W}{ap}}

First we consider position bias, based on the observation that a word at a given relative position 
  in the source tends to align to a word at a similiar relative position in the target, $\frac{i}{I} \approx \frac{j}{J}$ (IBM 2). Related, alignments tend to occur near the diagonal \cite{dyer-chahuneau-smith:2013:NAACL-HLT}, when considering the alignments as a binary $I \times J$ matrix (illustrated in Figure~\ref{fig:sentence-attentional}), where the cell at $(i,j)$ denotes whether an alignment exists between source word $i$  and target word $j$. 

We include a position bias through redefining the pre-normalised attention scalars $f_{ji}$ in Eq~\ref{eq:sent-fji} as:
\begin{align}
f_{ji}  = \vv^\top \tanh \big( &\Wae \vE_i + \Wah \vg_{j-1} + \nonumber\\
&  \Wap \psi(j, i, I) \big) \label{eq:sent-fji-positional} 
\end{align}
where the new component in the input is a simple feature function of
the positions in the source and target sentences and the source length, 
\[ \psi(j, i, I) = \bigg[ \log(1+j), \log(1+i), \log(1+I) \bigg]^{\top}  \]
and $\Wap \sizem{A}{3}$. We exclude the target length $J$ as this is unknown during
decoding, as a partial translation can have several (infinite)
different lengths. The use of the $\log(1 + \cdot)$ function is to 
avoid numerical instabilities from widely varying sentence lengths.
The non-linearity in Eq~\ref{eq:sent-fji-positional} allows for
complex functions of these inputs to be learned, such as relative
positions and approximate distance from the diagonal, as well as their
interactions with the other inputs (e.g., to learn that some words are
exceptional cases where a diagonal bias should not apply).

\subsection{Markov condition}

\newcommand{\Wam}{\msuper{W}{am}}
\newcommand{\Waf}{\msuper{W}{af}}

The HMM model of translation \cite{vogel96} is based on a Markov 
condition over alignment random variables, to allow the model to
learn local effects such as when $i \gets j$ is aligned then it is likely
that $i+1 \gets j+1$ or $i \gets j+1$. These correspond to
local diagonal alignments or one-to-many alignments, respectively.
In general, there are many correlations between the alignments of
a word and the word immediately to its left.

Markov conditioning can also be incorporated in 
a similar manner to positional bias, by augmenting the attentional input from
Eqs~\ref{eq:sent-fji}~and~\ref{eq:sent-fji-positional} to include:
\begin{equation}
f_{ji}  = \vv^\top \tanh \left( \ldots + \Wam \xi_1(\valpha_{j-1};i)\right) \label{eq:sent-fji-markov} 
\end{equation}
where  $\ldots$ abbreviates the $\ve_i$, $\vg_{j-1}$ and $\psi$ components from 
Eq~\ref{eq:sent-fji-positional}, and 
 $\xi_1(\valpha_{j-1})$ provides a fixed dimensional representation of
the attention state for the preceding word. 
It is not immediately obvious how to incorporate the previous attention vector
as $\valpha$ is dynamically sized to match the
source sentence length, thus using it directly would not
generalise over sentences of different lengths. For this reason, we
make a simplification by just considering local moves offset by $\pm k$ positions, that is, 
\begin{equation}
 \xi_1(\valpha_{j-1}; i) = \bigg[ \alpha_{j-1,i-k}, .., \alpha_{j-1,i},   .., \alpha_{j-1,i+k} \bigg ]^\top \nonumber
\label{eq:sent-xi}
\end{equation}
with $\Wam \sizem{A}{(2k+1)}$. 
Our approach is likely to capture the most important alignments patterns forming the backbone
of the alignment HMM, namely monotone, 1-to-many, and local inversions.


\subsection{Fertility}

Fertility is the propensity for a word to be translated as a consistent
number of words in the other language, e.g., \emph{Iseseisvusdeklaratsioon}
translates as \emph{(the) Declaration of Independence}.
Fertility is a central component in the IBM models
3--5 \cite{brown93}. Incorporating fertility into the
attentional model is a little more involved, and we present two techniques for
doing so. 

\paragraph{Local fertility}
First we consider a feature-based technique, which includes the following  features
{\small
\begin{equation}
 \xi_2(\valpha_{<j}; i) = \left[ \sum_{j'<j} \alpha_{j',i-k},.., \sum_{j'<j} \alpha_{j',i},  .., \sum_{j'<j} \alpha_{j',i+1} \right ]^\top \nonumber
\label{eq:sent-xi-fertility}
\end{equation}
}
and the corresponding feature weights, i.e., $\Waf \sizem{A}{(2k+1)}$. These sums
represent the total alignment score for the surrounding source words, similar 
to fertility in a traditional latent variable model, which is the
sum over binary alignment random variables. 
A word which already has several alignments can be excluded from participating in more alignments,
thus combating the garbage collection problem. Conversely words that
tend to need high fertility  
can be learned through the interactions between these features and the
word and context embeddings in Eq~\ref{eq:sent-fji-markov}.

\paragraph{Global fertility}
A second, more explicit, technique for incorporating fertility is to include 
this as a modelling constraint. Initially we considered a soft constraint based 
on the approach in \cite{icml2015_xuc15}, where an 
image captioning model was biased to attend to every pixel in the image 
exactly once. In our setting, the same idea can be applied through adding a
regularisation term to the training objective of the form $ \sum_i \left(1 - \sum_j \alpha_{j,i} \right)^2 $.
However this method is overly restrictive: enforcing that every word is used exactly once 
is not appropriate in translation where some words are likely to be dropped (e.g., determiners and other function words), while others
might need to be translated several times to produce a phrase in the target language.\footnote{
Modern decoders \cite{koehn2003statistical} often impose the restriction of each word being translated exactly once, however this is tempered by their use of phrases as translation units rather than words, which allow for higher fertility in contiguous translation chunks.
}
For this reason we develop an alternative method, based around a contextual fertility model,
$p(f_i | \vs, i) = \mathcal{N}\left( \mu(e_i), \sigma^2(e_i) \right) $
which scores the fertility of source word $i$, defined as $f_i = \sum_j \alpha_{j,i}$, using a normal
distribution\footnote{The normal distribution is deficient, as it has support for all scalar values, despite $f_i$ being bounded above and below ($0 \le f_i \le J$). This could be corrected by using a truncated normal, or various other choices of distribution.} parameterised by $\mu$ and $\sigma^2$, both positive scalar valued non-linear functions 
of the source word encoding $e_i$. This is incorporated into the training objective as an additional
additive term, $\sum_i \log p(f_i | \vs, i)$, for each training sentence.

This formulation allows for greater consistency in translation, 
through e.g., learning which words tend to be omitted from translation, or translate as several words.
Compared to the fertility model in IBM 3--5 \cite{brown93}, ours uses many fewer parameters through working over vector embeddings, and moreover, the BiRNN encoding of the source means that we learn context-dependent fertilities, which can be useful for dealing with fixed syntactic patterns or multi-word expressions.

\subsection{Bilingual Symmetry}
\label{sec:sent-alignment-symmetry}

So far we have considered a conditional model of the target given the
source, modelling $p(\vt | \vs)$. However it is well established for
latent variable translation models that the alignments improve if  $p(\vs | \vt)$ is also modelled
and the inferences of both directional models are combined -- evidenced
by the symmetrisation heuristics used in most decoders \cite{koehn2005iwslt},
and also by explicit joint agreement training objectives \cite{liang2006alignment,ganchev-gracca-taskar:2008:ACLMain}.
The rationale is that both models make somewhat independent errors, so an ensemble
stands to gain from variance reduction.


%

\begin{figure}[t]
\centering
\includegraphics[width=0.8\columnwidth]{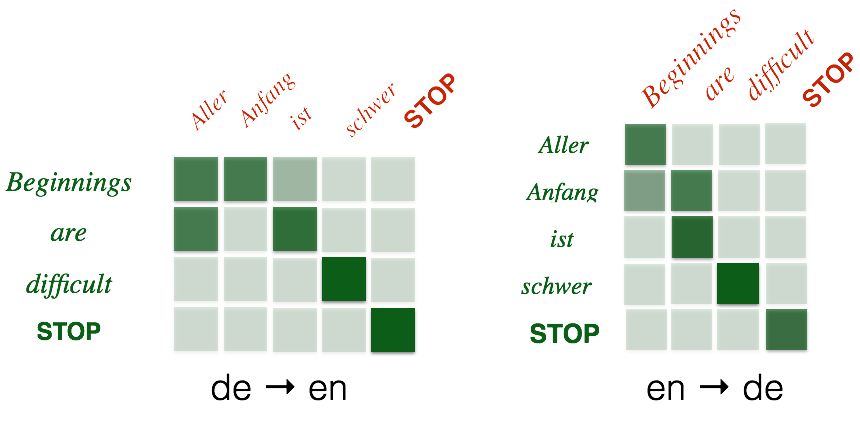}
\caption{Symmetric training with trace bonus, computed as matrix multiplication,  $-\operatorname{tr}(\valpha^{s\leftarrow t} \valpha^{s\rightarrow t ~ \top})$. Dark shading indicates higher values. }
\label{fig:btrace}
\end{figure}

We propose a method for joint training of two directional models as
pictured in Figure~\ref{fig:btrace}.  Training twinned models involves optimising
\mbox{$\mathcal{L}  = -\log p(\vt | \vs) - \log p(\vs | \vt) + \gamma B $} 
where, as before, we consider only a single sentence pair, for simplicity of notation.
This corresponds to a pseudo-likelihood objective, with the $B$ linking
the two models.\footnote{We could share some parameters, e.g., the word
  embedding matrices, however we found this didn't make much difference
  versus using disjoint parameter sets. We set $\gamma=1$ herein.} 
The $B$ component considers the alignment (attention) matrices,
$\valpha^{s\rightarrow t} \sizem{J}{I}$ and $\valpha^{t \leftarrow s} \sizem{I}{J}$, and attempts to make these close to one another for both
translation directions (see Fig.~\ref{fig:btrace}). To achieve this,
we use a `trace bonus', inspired by \cite{levinboim15}, formulated as
\begin{align*}
B = -\operatorname{tr}(\valpha^{s\leftarrow t~\top} \valpha^{s\rightarrow t}) & = \sum_j \sum_i \alpha^{s\leftarrow t}_{i,j} \alpha^{s\rightarrow t}_{j,i}\, .
\end{align*}
As the alignment cells are normalised using the $\operatorname{softmax}$ and thus take
values in [0,1],  the trace term is bounded above by $\min(I,J)$ which
occurs when the two alignment matrices are transposes of each other, representing perfect one-to-one alignments in both directions


\section{Experiments}

\begin{table}
\centering
\begin{tabular}{|c||S[table-format = 4.0]S[table-format = 4.0]|S[table-format = 3.2]S[table-format = 3.2]|}
\hline
\textbf{lang-pair} &  \multicolumn{2}{c|}{\textbf{\# tokens (K)}}  & \multicolumn{2}{c|}{\textbf{\# types (K)}} \\
\hline \hline
Zh-En &  422 & 454 & 3.44 & 3.12 \\
Ru-En &  1639 & 1809 & 145 & 65 \\
Et-En &  1411 & 1857 & 90 & 25 \\
Ro-En &  1782 & 1806 & 39 & 24 \\
\hline
\end{tabular}
\caption{Statistics of the training sets, showing in each cell the count for the source language (left) and target language (right).}

\label{datasets_tab}
\end{table}

\paragraph{Datasets.} We conducted our experiments with four language pairs, translating between English $\leftrightarrow$ Romanian, Estonian, Russian and Chinese. 
These languages were chosen to represent a range of translation difficulties, including languages with significant morphological complexity (Estonian, Russian).
We focus on a (simulated) low resource setting, where only a limited amount of training data is available.
This serves to demonstrate the robustness and generalisation of our model on sparse data -- something that 
has not yet been established for neural models with millions of parameters with vast potential for over-fitting.

Table \ref{datasets_tab} shows the statistics of the training sets.\footnote{For all datasets words were thresholded for training frequency $\ge 5$, with uncommon training and unseen testing words replaced by an \unk symbol.}
For Chinese-English, the data comes from the BTEC corpus, where the number of training sentence pairs is 44,016. 
We used `devset1\_2' and `devset\_3' as the development and test sets, respectively, and in both cases used only the first reference for evaluation.  
For other language pairs, the data come from the Europarl corpus \cite{koehn2005epc}, where we used 100K sentence pairs for training, and 3K for development and 2K for testing.\footnote{The first 100K sentence pairs were used for training, while the development and test were drawn from the last 100K sentence pairs, taking the first 2K for testing and the last 3K for development.}

\paragraph{Models and Baselines.} We have implemented our neural translation model with linguistic features 
in C++ using the CNN library.\footnote{\url{https://github.com/clab/cnn/}} 
We compared our proposed model against our implementations of 
the attentional model \cite{bahdanau2015neural} and
encoder-decoder architecture \cite{sutskever2014sequence}. 
As the baseline, we used a state-of-the-art phrase-based statistical machine translation model 
built using Moses \cite{koehn2007moses} with the standard
features: relative-frequency and lexical translation model probabilities in both directions;
distortion model; language model and word count. 
We used KenLM  \cite{Heafield-kenlm}  to create  3-gram
language models with Kneser-Ney smoothing on the target side of the bilingual training corpora.

\paragraph{Evaluation Measures.} 

Following previous work \cite{kalchbrenner13emnlp,sutskever2014sequence,bahdanau2015neural,neubig15wat}, 
we evaluated all neural models using test set perplexities and in a re-ranking setting, using BLEU \cite{Papineni:2002:BMA:1073083.1073135} measure.
For re-ranking, we generated 100-best translations using the baseline phrase-based model, 
to which we added log probability features from our neural models alongside  
the features of the underlying phrase-based model.

\subsection{Analysis of Alignment Biases}

\begin{table}
\centering
\sisetup{
round-mode = places,
round-precision = 2,
detect-weight,
table-format = 2.2
}%
\begin{tabular}{|l||S|S[round-precision = 1,table-format=2.1]|}
\hline 
\bf configuration & 
{ \bf test} & {\bf \#param (M)} \\
\hline \hline
  Sutskever encdec & 5.34871 & 8.687665 \\
\hline
 Attentional & 4.76879 & 14.977841 \\
 +align  & 4.56259 & 14.980145 \\
 +align+glofer &  5.20311 & 15.505971 \\
 +align+glofer-pre & 4.31061 & 15.505971 \\
 +align+sym & 4.4404 & 30.121885 \\
 +align+sym+glofer-pre & 4.432 & 31.173537 \\
\hline
\end{tabular}
\caption{Perplexity results for attentional model variants evaluated on BTEC zh$\rightarrow$en, and number of model parameters (in millions).}
\label{tab:sent-btec-pplx}
\end{table}

\begin{figure}[t]
\centering
\includegraphics[width=0.8\columnwidth]{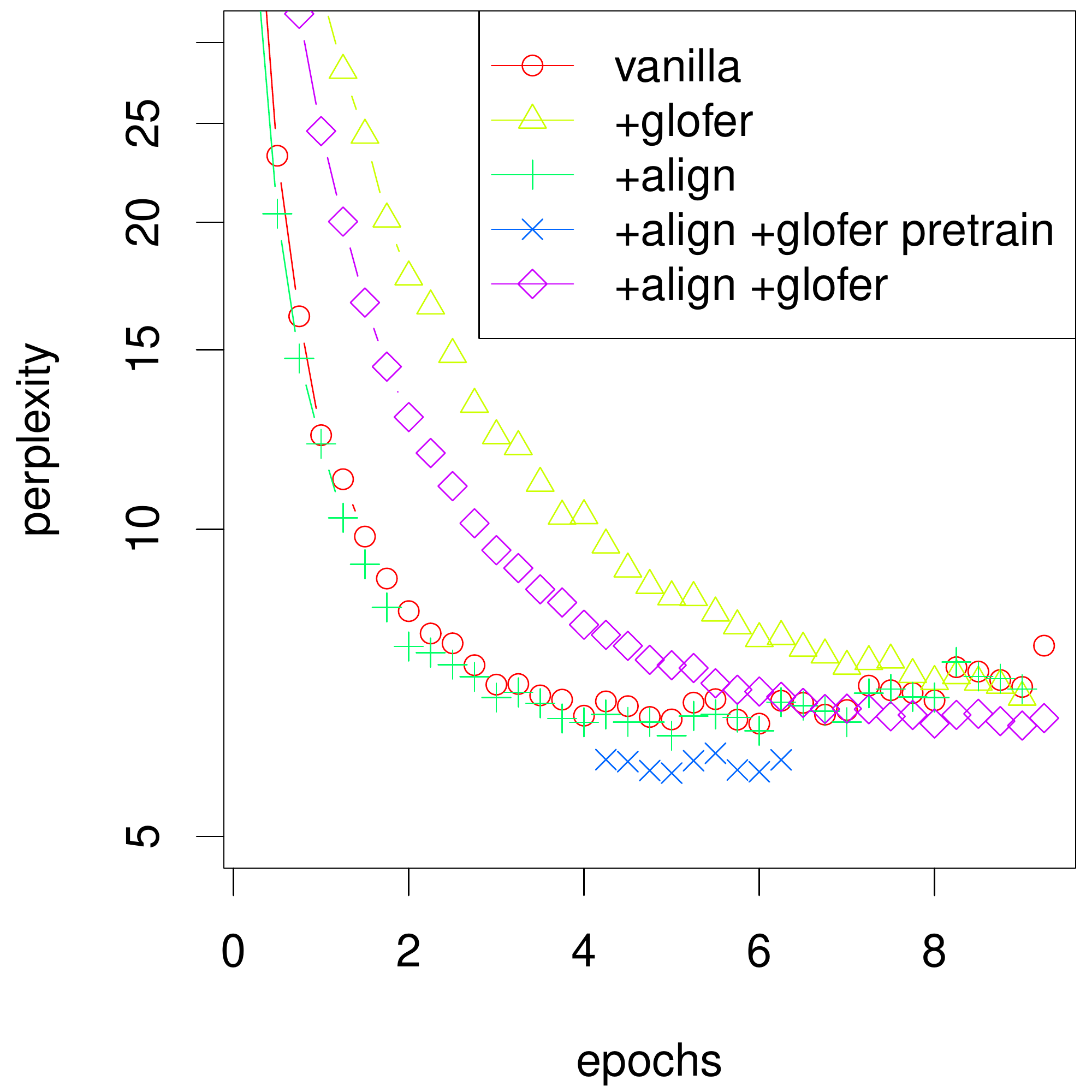}
\caption{Perplexity with training epochs on ro-en translation, comparing several model variants.}
\label{fig:ro-en-abelation}
\end{figure}

\begin{figure*}[!t]
\centering
\includegraphics[width=\textwidth]{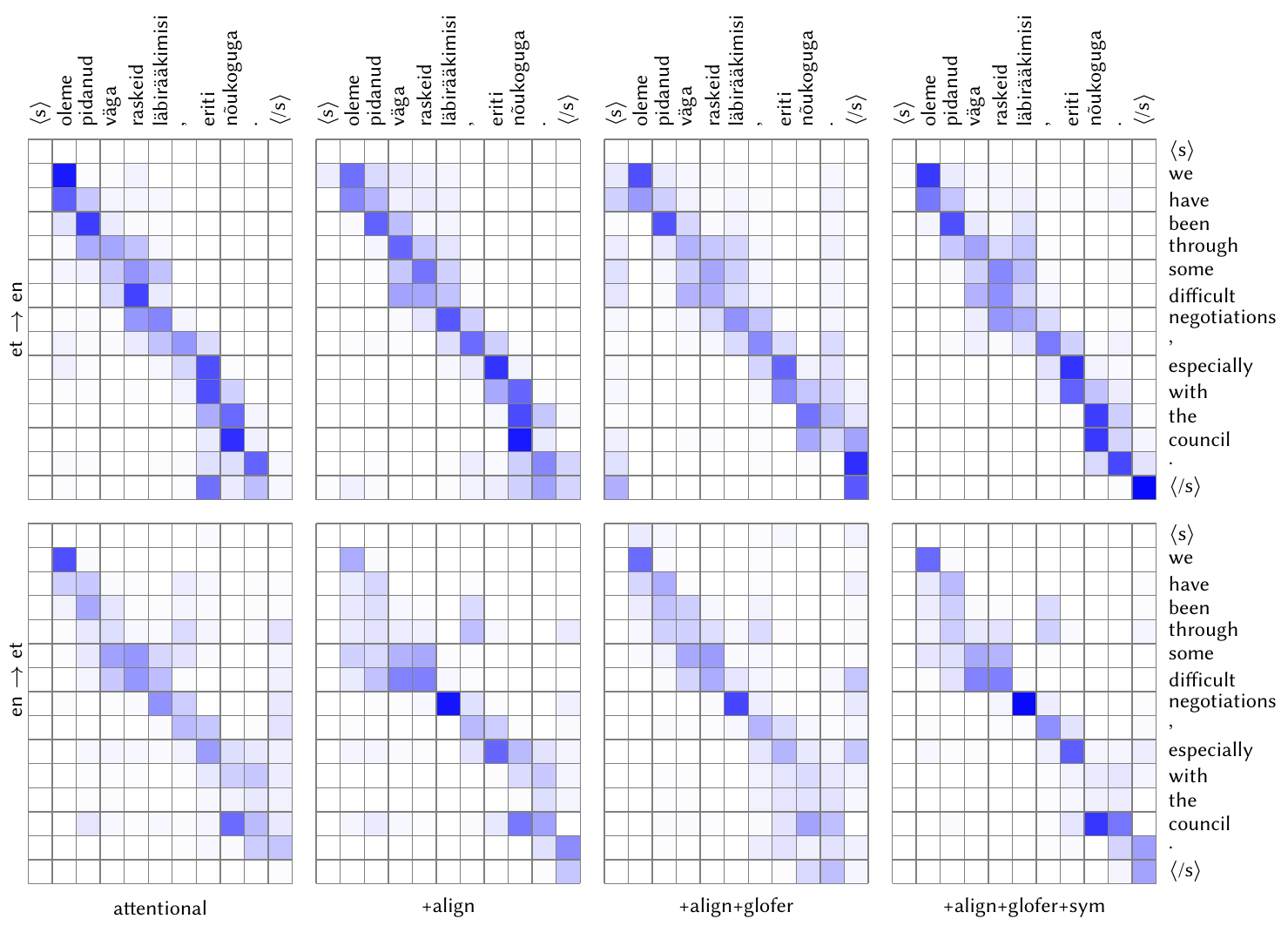}
\caption{Example development sentence, showing the inferred attention matrix for various models for Et~$\leftrightarrow$~En. Rows correspond to the translation direction and columns correspond to different models: attentional, with
alignment features (+align), global fertility (+glofer), and symmetric joint training (+sym). Darker shades denote higher values and white denotes zero.}
\label{fig:agreement}
\end{figure*}

We start by investigating the effect of various linguistic constraints, 
described in Section \ref{sec:sent-alignment-structure}, on the attentional model.
Table~\ref{tab:sent-btec-pplx} presents the perplexity of trained models
for Chinese$\rightarrow$English translation.
%
%
For comparison, we report the results of an encoder-decoder-based  neural translation  model \cite{sutskever2014sequence} as the baseline. 
%
All other results are for the attentional model with a single-layer LSTM as encoder and two-layer LSTM as decoder, 
using 512 embedding, 512 hidden, and 256 alignment dimensions.
For each model, we also report the number of its parameters. 
Models are trained using stochastic gradient, allowing up to 20 epochs. 
For each model the best perplexity on the held-out development set is reported, which was achieved in 5-10 epochs for most cases.

As expected, the vanilla attentional model greatly improves over encoder-decoder (perplexity of 4.77 vs.\@ 5.35), 
clearly making good use of the additional context.
%
%
Adding the combined positional bias, local fertility, and Markov structure (denoted by +align) 
further decreases the perplexity to 4.56. 
Adding the global fertility (+glofer) is detrimental, however, increases perplexity to 5.20. 
Interestingly, global fertility does helps to reduce the perplexity (to 4.31) when using with pre-training setting (+align+glofer-pre). In this case, it is refining an already excellent model from which reliable global fertility estimates can be obtained.
This finding is consistent with the other languages, see Figure~\ref{fig:ro-en-abelation} which shows typical learning curves of different variants of the attentional model. 
Note that when global fertility is added to the vanilla attentional model with alignment features, it significantly slows down training as it limits exploration in early training iterations, however it does bring a sizeable win when used to fine-tune a pre-trained model.
%
%
%
Finally, the bilingual symmetry also helps to reduce the perplexity scores when used with the alignment features, however, does not combine well with global fertility (+align+sym+glofer-pre).
This is perhaps an unsurprising result as both methods impose a often-times similar regularising effect over the attention matrix.

%
%
%
%

Figure \ref{fig:agreement} illustrates the different attention matrices inferred by the various model variants. 
Note the difference between the base attentional model and its variant with alignment features (`+align'), 
where more weight is assigned to diagonal and 1-to-many alignments. 
Global fertility pushes more attention to the sentinel symbols \sos and \eos. 
Determiners and prepositions in English show much lower fertility than nouns, 
while Estonian nouns have even higher fertility.
This accords with Estonian morphology, wherein nouns are inflected with rich case 
marking, e.g., \emph{n\~{o}ukoguga} has the cogitative \emph{-ga} suffix, meaning `with', 
and thus translates as several English words (\emph{with the council}).
The right-most column corresponds to joint symmetric training, with many more 
confident attention values especially for consistent 1-to-many alignments 
(\emph{difficult} in English and \emph{raskeid} in Estonian, an adjective in partitive case meaning \emph{some difficult}).

%

\subsection{Full Results}

The perplexity results of the neural models for the two translation 
directions across the four language pairs are presented in Table~\ref{res:perplexity}.a 
and~\ref{res:perplexity}.b. 
In all cases, our work achieves lower perplexities compared to 
the vanilla attentional model and the encoder-decoder architecture, 
owing to the linguistic constraints. 

Table \ref{res:bleu:rerank} presents the BLEU scores for the re-ranking setting for the 
translating into English from our four languages.
We compare re-ranking settings using the log probabilities produced by our model as additional features 
vs.\@ using log probabilities from the vanilla attentional model and the encoder-decoder.  
The re-rankers based on our model are significantly better than the rest 
for Chinese and Estonian, and on par with the other for Russian and Romanian$\rightarrow$English.
In all cases our model has performance at least 1 BLEU point better than the baseline phrase-based system.
It is worth noting that for Chinese-English, our re-ranker leads to an increase of almost 3 points in the BLEU score using an \emph{ensemble} of neural models with different 
configurations.\footnote{We use the outputs of 6--12 models trained in both directions, using different alignment and fertility options, and using a smaller dimensionality than earlier (100 embedding, 100 hidden and 50 attention dimensions).}

\begin{table}
\centering
\begin{tabular}{c}
{\small
\sisetup{
round-mode = figures,
round-precision = 3,
detect-weight,
table-format = 2.2
}%
\robustify\bfseries
\begin{tabular}{|l||S[detect-weight,table-format = 3.3]SSS|}
\hline
Lang. Pair & {Zh-En} & {Ru-En} & {Et-En} & {Ro-En} \\
\hline \hline
Enc-Dec & 5.34871 & 61.9471 & 18.2496 & 10.2685 \\
Attentional & 4.76879 & 41.7227 & 12.7635 & 6.62272 \\
Our Work & \bfseries 4.31061 & \bfseries 39.8725 & \bfseries 11.82 & \bfseries 5.89019 \\
\hline
\end{tabular} 
} \\
(a) \\
{\small
\sisetup{
round-mode = figures,
round-precision = 3,
detect-weight,
table-format = 2.2
}%
\robustify\bfseries
\begin{tabular}{|l||SSSS|}
\hline
Lang. Pair & {En-Zh} & {En-Ru} & {En-Et} & {En-Ro} \\
\hline \hline
Enc-Dec & 8.5965 & 67.31 & 31.4038 & 11.5043 \\
Attentional & 7.48505 & 43.0271 & 19.4009 & 7.29634 \\
Our Work & \bfseries 6.24009 & \bfseries 40.6302 & \bfseries 16.9558 & \bfseries 6.34599 \\
\hline
\end{tabular} 
} \\
(b) \\
\end{tabular}
\caption{Perplexity on the test sets for the two translation directions. Our work includes: bidirectional LSTM attentional model combined with positional bias, Markov, local fertility, and global fertility (pre-trained setting).}
\label{res:perplexity}
\end{table}


{\small

\begin{table}
\centering
\sisetup{
round-mode = figures,
round-precision = 4,
detect-weight,
table-format = 2.2
}%
\robustify\bfseries
\resizebox{\columnwidth}{!}{
\begin{tabular}{|l||SSSS|}
\hline
Lang. Pair & {Zh-En} & {Ru-En} & {Et-En} & {Ro-En} \\
\hline \hline
Phrase-based  & 40.63 & 18.7 & 31.99&  45.21\\
\hline
 Enc-Dec  & 40.41 & 18.83 & 32.20 & 45.36 \\
 Attentional   & 41.16 \textsuperscript{$\clubsuit$}& \bfseries 19.79 & 32.78 & 46.83 \\

Our Work  & \bfseries 44.14 \textsuperscript{$\clubsuit\spadesuit$} & 19.73 & \bfseries 33.26 \textsuperscript{$\spadesuit$} & \bfseries 46.88 \\
\hline
\end{tabular}}
\caption{BLEU scores  on the test sets for re-ranking.
\textbf{bold:} Best performance, $^{\spadesuit}$: Significantly better than Attentional,  $^{\clubsuit}$: Using ensemble of models. 
} 
\label{res:bleu:rerank}
\end{table}
}

\section{Related Work}

%
\newcite{kalchbrenner13emnlp}
were the first to propose a full neural model of translation, using a convolutional network as the source encoder, followed by an RNN decoder to generate the target translation.
This was extended in \newcite{sutskever2014sequence}, who replaced the source encoder with an RNN using a Long Short-Term Memory (LSTM), and \newcite{bahdanau2015neural} who introduced the notion of ``attention'' to the model, whereby the source context can dynamically change during the decoding process to attend to the most relevant parts of the source sentence
\newcite{luong-pham-manning:2015:EMNLP} refined the attention mechanism to be more local, by 
constraining attention to a text span, whose words' representations are averaged.
To leverage the attention history, \cite{luong-pham-manning:2015:EMNLP}  made use
of the attention vector of the previous position when generating the attention vector for the next position, similar in spirit to our method for incorporating alignment structural biases.
Concurrent with our work, \newcite{chengetal2015} proposed a similar agreement-based
joint training for bidirectional attention-based neural machine translation, and showed significant improvement in the BLEU score for the large data French$\leftrightarrow$English translation.

\section{Conclusion}

We have shown that the attentional model of translation does not capture many 
well known properties of traditional word-based translation models, and proposed
several ways of imposing these as structural biases on the model. We show 
improvements across several challenging language pairs in a low-resource setting,
both in perplexity and re-ranking evaluations. In future work we intend to
investigate the model performance on larger datasets, and incorporate further linguistic information such as morphological representations.

\ifnaaclfinal
\subsection*{Acknowledgements}

The work reported here was started at JSALT 2015 in UW, Seattle, and was supported by JHU via grants from NSF (IIS), DARPA (LORELEI), Google, Microsoft, Amazon, Mitsubishi Electric, and MERL. Dr Cohn was supported by the ARC (Future Fellowship).
\fi


\bibliographystyle{naaclhlt2016}
\bibliography{cite-strings,sentence,cite-definitions}

\end{document}